\pdfoutput=1
\documentclass[10pt,twocolumn,letterpaper]{article}

\usepackage[pagenumbers]{cvpr} %

\definecolor{cvprblue}{rgb}{0.21,0.49,0.74}
\usepackage[pagebackref,breaklinks,colorlinks,allcolors=cvprblue]{hyperref}
\usepackage{multirow}
\usepackage{adjustbox} %
\usepackage{dblfloatfix} %
\usepackage{placeins}
\usepackage[numbers]{natbib}

\def\paperID{14527} %
\def\confName{CVPR}
\def\confYear{2025}

\title{Gradient-Weighted Feature Back-Projection: A Fast Alternative to Feature Distillation in 3D Gaussian Splatting}

\author{Joji Joseph\\
Indian Institute of Science\\
{\tt\small jojijoseph@iisc.ac.in}
\and
Bharadwaj Amrutur\\
Indian Institute of Science\\
{\tt\small amrutur@iisc.ac.in}
\and
Shalabh Bhatnagar\\
Indian Institute of Science\\
{\tt\small shalabh@iisc.ac.in}
}

\begin{document}
\maketitle
\begin{abstract}
We introduce a training-free method for feature field rendering in Gaussian splatting. Our approach back-projects 2D features into pre-trained 3D Gaussians, using a weighted sum based on each Gaussian's influence in the final rendering. While most training-based feature field rendering methods excel at 2D segmentation but perform poorly at 3D segmentation without post-processing, our method achieves high-quality results in both 2D and 3D segmentation. Experimental results demonstrate that our approach is fast, scalable, and offers performance comparable to training-based methods. Project page: \url{https://jojijoseph.github.io/3dgs-backprojection/}
\end{abstract}
    
\section{Introduction}
\label{sec:intro}

3D Gaussian Splatting (3DGS) [3] is a novel-view synthesis technique that uses 3D Gaussians as rendering primitives, each defined by its mean position and variance. Additionally, these Gaussians carry payloads such as opacity and anisotropic color, enabling photorealistic rendering.

Generating intermediate feature maps directly, rather than passing RGB renderings to a foundation model, can be advantageous for tasks like segmentation. However, training Gaussian splatting models to render feature maps directly is computationally intensive, especially given the high dimensionality of feature space.

Moreover, segmentation queries are more effective when performed on rendered features than on features assigned to each Gaussian, as only the surface Gaussians typically align with the feature map. Since rendered features result in a weighted sum, Gaussians deeper within the scene may not accurately correspond to features in the feature map, complicating accurate 3D segmentation.

Given these drawbacks in training time and segmentation accuracy, alternative methods are essential. One approach is to aggregate 2D features into Gaussians that influence the regions where the 2D features are projected. In other words, the Gaussian features are computed as a weighted sum of the 2D features, with weights proportional to the influence of each Gaussian on the corresponding region in the 2D feature map.

Our main contributions are as follows:

\begin{itemize} \item We introduce a training-free approach that projects 2D features onto 3D Gaussians, providing a fast and scalable alternative to traditional feature field distillation. \item We demonstrate that our method achieves comparable or superior performance to feature field distillation methods that rely on extensive training, particularly in producing clean 3D segmentations. \item Our method effectively supports both rendered feature space and 3D feature space, enabling seamless querying for downstream applications such as 3D object manipulation and real-time scene understanding. \end{itemize}

\section{Related Works}
\label{sec:related}

Neural Radiance Fields (NeRF) \cite{mildenhall2020nerf} has emerged as a leading approach for synthesizing novel views from sparse image inputs. By leveraging a neural network, NeRF captures a volumetric representation of a scene, modeling the color and density at every 3D location to generate highly realistic views from arbitrary perspectives. However, the implicit nature of NeRF’s scene representation poses challenges for tasks like object modification or rearrangement, as such operations typically demand retraining the entire network.

In contrast, 3D Gaussian Splatting \cite{kerbl3Dgaussians} provides an explicit representation of 3D scenes, using 3D Gaussians as the primary rendering primitives. Attributes like color, opacity, and orientation characterize these Gaussians. This explicit structure allows for direct manipulation of Gaussians and their associated parameters, enabling efficient object rearrangement and editing. Such flexibility makes 3D Gaussian Splatting well-suited for interactive applications, including object editing, augmented reality, digital twins, and robotics.

An intuitive progression from radiance field rendering is feature field rendering, which incorporates additional feature embeddings to enrich the representation. Recent studies, such as \cite{zhou2024feature, qin2023langsplat, shi2023language}, have advanced this concept for tasks like segmentation and semantic querying.

Feature-3DGS \cite{zhou2024feature} focuses on training high-dimensional feature embeddings, while LangSplat \cite{qin2023langsplat} prioritizes compressed, low-dimensional features. Both methods demonstrate strong performance in 2D segmentation of rendered outputs but face significant challenges with 3D object segmentation. Feature-3DGS attempts 3D segmentation by matching language embeddings with Gaussian feature embeddings. However, this approach often falls short because the features of individual Gaussians do not directly map to the final rendered feature, which results from the weighted sum of contributions from multiple Gaussians (see Equation~\ref{eq:3dgs-alpha-blending}). This inherent mismatch hinders reliable 3D segmentation. In contrast, our method overcomes this limitation by directly leveraging gradient information, enabling more accurate and effective 3D segmentation.

Another class of methods tackles object segmentation from visual prompts such as points and masks\cite{cen2023saga, hu2024sagdboundaryenhancedsegment3d, flashsplat, joji2024gradient}. In SAGD \cite{hu2024sagdboundaryenhancedsegment3d}, a binary voting system is employed, while methods in \cite{flashsplat} and \cite{joji2024gradient} use similar influence-based voting approaches. FlashSplat \cite{flashsplat} formulates segmentation as an optimization problem, whereas \cite{joji2024gradient} leverages inference-time gradient backpropagation. Our approach also utilizes inference-time backpropagation for feature back-projection, enabling robust feature representation across 3D space.

\section{Method}
\label{sec:method}

In this section, we present the feature back-projection equation and describe four direct use cases — 3D segmentation, affordance transfer, and identity encoding—that do not require post-processing other than similarity search. Notably, we perform these use cases directly in 3D space rather than in 2D image space, as is common in similar works.

\subsection{Feature Back-Projection}
Consider the color $C$ of a pixel at $(x, y)$ in a 3DGS rendering,
\begin{align}
    C(x,y) &= \sum_{n\leq N} c_n \alpha_n(x,y) \prod_{m  < n} ( 1 - \alpha_m(x,y)) \\ &=\sum_{n\leq N} c_n \alpha_n(x,y) T_n(x,y)
    \label{eq:3dgs-alpha-blending}
\end{align}

Where $N$ is the total number of Gaussians, each indexed by its sorted position, $c_n$ is the color associated with the $n$th Gaussian, $\alpha_n(x,y)$ is the opacity of the $n$th Gaussian at $(x,y)$ adjusted with exponential falloff, and \mbox{$T_n=\prod_{m  < n} ( 1 - \alpha_m(x,y))$} is the transmittance of $n$th Gaussian at $(x,y)$.

From this equation, it’s clear that the rendered color is a weighted sum of colors of individual Gaussians. Moreover, the weight is opacity multiplied by transmittance.

Taking the derivative with respect to color of $k$th Gaussian $c_k$,
\begin{align}
    \frac{ \partial{C(x,y)}} {\partial{c_k}} = \alpha_k(x,y) T_k(x,y)
\end{align}

This gradient is equivalent to the weight of each Gaussian in a pixel rendering. This insight enables us to leverage inference-time gradient backpropagation to compute feature back-projections based on the Gaussian’s influence efficiently.

Given this gradient-based weighting, we can define the feature back-projection equation as follows:

\begin{equation}
    \textbf{f}_k = \frac{\sum_{(x,y,n)} \textbf{F}_{2D}(x,y,n)\alpha_k(x,y,n)T_k(x,y,n)}{\sum_{(x,y,n)} \alpha_k(x,y,n)T_k(x,y,n)}
    \label{eq:feature-back-projection}
\end{equation}

We call this equation the expected feature back-projection equation or simply the back-projection equation.

Where $\textbf{f}_k$ is the feature of $k$th Gaussian. $\textbf{F}_{2D}(x,y,n)$ is the feature at $(x,y)$ in the $n$th viewpoint. This formulation allows for efficient aggregation of features across viewpoints, weighted by opacity and transmittance, resulting in an accurate feature back-projection that reflects the Gaussian’s contribution to the final rendered scene.

When we remove the denominator from this equation, it becomes accumulated back-projection.

\begin{equation}
    \textbf{f}_k = {\sum_{(x,y,n)} \textbf{F}_{2D}(x,y,n)\alpha_k(x,y,n)T_k(x,y,n)}
    \label{eq:accumulated-feature-back-projection}
\end{equation}

If we replace the feature $\textbf{F}_{2D}(x,y,n)$ with a function indicating binary function that indicates the presence of a mask this simply becomes the vote using masked gradients as described in \cite{joji2024gradient}.

If we do Euclidian normalization on $\textbf{f}_k$ after calculation, both equations \ref{eq:feature-back-projection}, \ref{eq:accumulated-feature-back-projection} become equivalent.

\subsection{3D Segmentation}
\label{sec:method-segmentation}

Once the feature back-projection is complete, we obtain a set of feature vectors of dimension $D$, \( \{\textbf{f}_k\} \in \mathbb{R}^D \) in the projected feature space. Let \( \textbf{q} \in \mathbb{R}^D\) represent the embedding extracted from the language query (or a query from another modality). The scalar value \( \mathrm{sim}(\mathbf{f}_k, \mathbf{q}) \), where \( \mathrm{sim} \) denotes a similarity function (typically cosine similarity), measures the relevance of each Gaussian \( g_k \) to the query. %

To perform segmentation, we apply a threshold on $ \mathrm{sim}(\mathbf{f}_k, \mathbf{q}) $, allowing us to isolate Gaussians that correspond closely to the specified query. Formally, the set of segmented Gaussians $G$ can be defined as:
\begin{equation}
G = \{ {g}_k \mid \mathrm{sim}(\mathbf{f}_k, \mathbf{q}) > \theta \},
\end{equation}
where \( \theta \) is a user-defined threshold.

Segmentation can also be performed in 2D. The equation~\ref{eq:2d-segmentation} represents the 2D mask by querying over rendered 2D features.

\begin{equation}
    M(x, y) = 
    \begin{cases}
        1, & \text{if } \mathrm{sim}(\mathbf{f}_k, \mathbf{q}) > \theta \\
        0, & \text{otherwise}
    \end{cases}
    \label{eq:2d-segmentation}
\end{equation}

Here, $M(x,y)$ is a binary mask that identifies regions in 2D space where the similarity to the query exceeds the threshold.

\subsection{Affordance Transfer}

Affordance transfer\cite{Hadjivelichkov2022affcorrs} is the process of transferring annotated regions from source images to a target object. Although the source and target objects may differ in appearance, they belong to the same category, enabling the transfer of meaningful information across instances. This technique is a particular case of few-shot segmentation, commonly applied in robotics, where annotations of irregularly shaped regions are transferred to a target object. This transfer allows a robot to manipulate the target object using knowledge from the annotated regions.

Our approach follows a straightforward pipeline. First, we transfer DINOv2\cite{oquab2023dinov2, darcet2023vitneedreg} features to the target scenes via feature back-projection. Then, we identify the closest matching source annotation for each Gaussian in the target scene by performing a k-nearest neighbors (kNN) classification on the feature space. Here, the kNN classification finds source annotations most similar to the Gaussian features, ensuring that the transferred affordances correspond closely to the target object’s structure.

\subsection{Identity Encoding}

When annotations are available for different objects within a scene, we can encode each object with a unique vector and back-project these vectors into the 3D scene. Even if annotations are missing for some views, this method can generalize across the entire 3D scene given sufficient annotated views.

We implement two types of identity encoding, each with unique strengths:

\textbf{Orthogonal Encoding}: In this approach, we assign procedurally generated orthogonal vectors to represent each object uniquely, where each vector direction is associated with a distinct object. The orthogonality of these vectors ensures that each object has a distinct, non-overlapping identity in the feature space, maximizing separation and simplifying the segmentation task. This back-projection operation can be viewed as an extension of segmentation with masked gradients \cite{joji2024gradient}, enabling the simultaneous encoding of multiple classes within a single scene. While effective, orthogonal encoding becomes challenging when representing a large number of objects, as the dimensionality of the feature space limits orthogonal vectors.

\textbf{Contrastive Encoding}: For cases where we need to encode many objects, we use non-orthogonal vectors made sufficiently distinct through contrastive learning. Contrastive learning maximizes the distance between feature vectors of different objects by pulling representations of different objects apart and bringing representations of the same object closer together. This method allows us to encode more objects than possible with strictly orthogonal vectors while still achieving clear separation between object identities in feature space. Although the separation is not as strict as with orthogonal encoding, contrastive learning provides a scalable and effective alternative, maintaining discriminative power even with far more objects than the embedding dimension.

The loss function for training the Identity Encoder-Decoder model is defined as:
\begin{align}
    \mathcal{L} &= \mathcal{L}_{\text{classification}} + \mathcal{L}_{\text{orthogonality}}, \\
    \mathcal{L}_{\text{classification}} &= \text{CrossEntropy}(\hat{y}, y), \\
    \mathcal{L}_{\text{orthogonality}} &= \| \mathbf{E} \mathbf{E}^\top - \mathbf{I} \|_F.
\end{align}

Where \( \hat{y} \) is the model’s predicted class probability and \( y \) is the true class label. \( \mathbf{E} \) is an embedding matrix of size \( N \times D \), where \( N \) is the number of classes and \( D \) is the embedding dimension. \( \mathbf{I} \) is the identity matrix and \( \| \cdot \|_F \) denotes the Frobenius norm.

\section{Experiments and Results}
\label{sec:experiments}

\subsection{Setup}
We use gsplat\cite{ye2024gsplatopensourcelibrarygaussian} as our rasterizer. All experiments are performed on NIVIDA A6000 GPU.

We use LSeg\cite{li2022languagedriven}, DINOv2\cite{oquab2023dinov2,darcet2023vitneedreg} features for our experiments. We also show we can encode objects using procedurally generated features as in orthogonal encoding and latent features generated by contrastive encoding.

There are cases where the numerator in the back-projection equation(\ref{eq:feature-back-projection}) is zero since there is no influence of Gaussian in the screen. In those cases, we can prune out the Gaussians with zero numerators before proceeding with inference.

\subsection{3D Segmentation}
\label{sec:exp-segmentation}

In this section, we use LSeg\cite{li2022languagedriven} features for our segmentation. We compare it qualitatively with Feature 3DGS. We use a custom implementation of Feature 3DGS based on gsplat \cite{ye2024gsplatopensourcelibrarygaussian} to prevent excessive RAM usage. We train Feature 3DGS versions for 7000 iterations. The training process takes around 20-30 minutes on our system, while the same number of iterations of Vannila 3DGS, which doesn't support feature field training, takes only 2 minutes 30 seconds. That is 10x faster compared to using Feature 3DGS.

After back-projection, we do segmentation as follows,

Let $\textbf{Q}=\{\textbf{q}_i\}$ be the set of prompts. Let $\textbf{q}_0$ be the category to segment, $\textbf{q}_i, i \neq 0$ are negative prompts. We use the last prompt as `other'.

The 3D mask is found by taking cosine similarity with Gaussian with both positive and negative prompts. Then, take the Gaussians, where the category prompt gives the highest cosine similarity as the 3D mask. Additionally, we can use a threshold over the cosine similarity with the category prompt.

\begin{figure}[h!]
    \centering
    \begin{tabular}{cc}
        \toprule
        \textbf{2D Segmentation (F3DGS)} & \textbf{2D Segmentation (Ours)} \\
        \midrule
        \adjustbox{valign=c}{\includegraphics[width=0.48\linewidth]{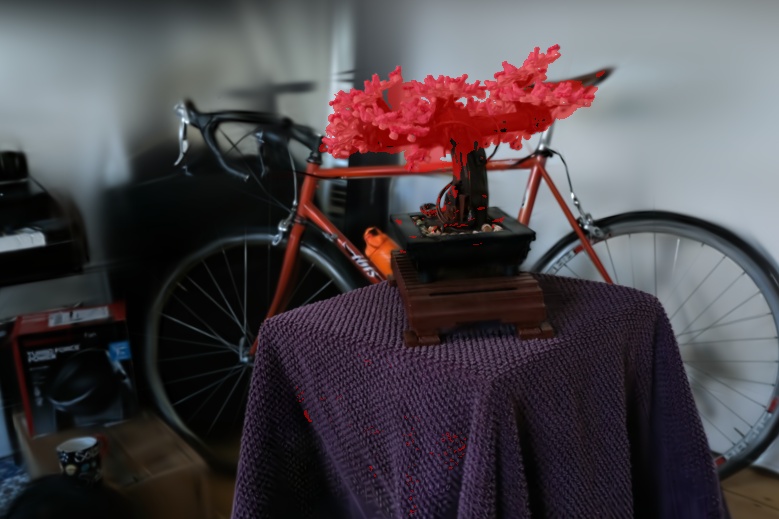}} &
        \adjustbox{valign=c}{\includegraphics[width=0.48\linewidth]{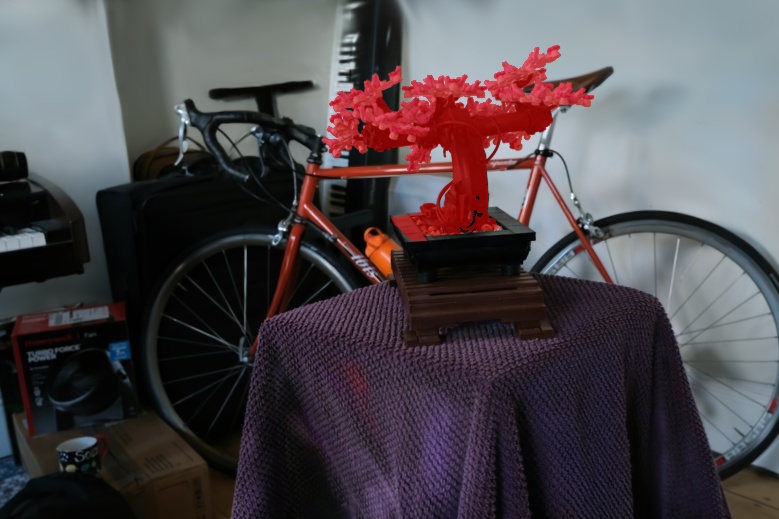}} \\
        \multicolumn{2}{c}{Prompt: Plant, Negative Prompt: Other}\\
        \multicolumn{2}{c}{}\\
        \adjustbox{valign=c}{\includegraphics[width=0.48\linewidth]{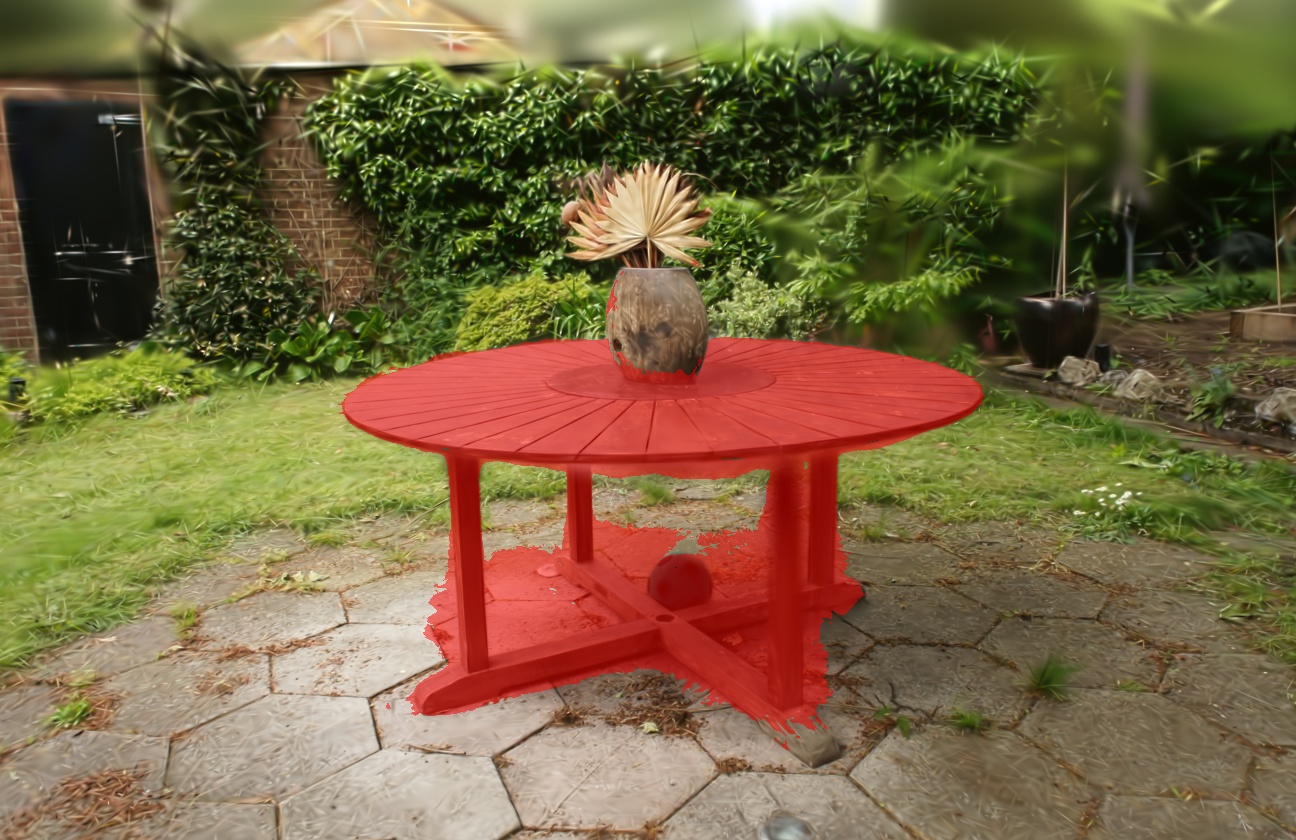}} &
        \adjustbox{valign=c}{\includegraphics[width=0.48\linewidth]{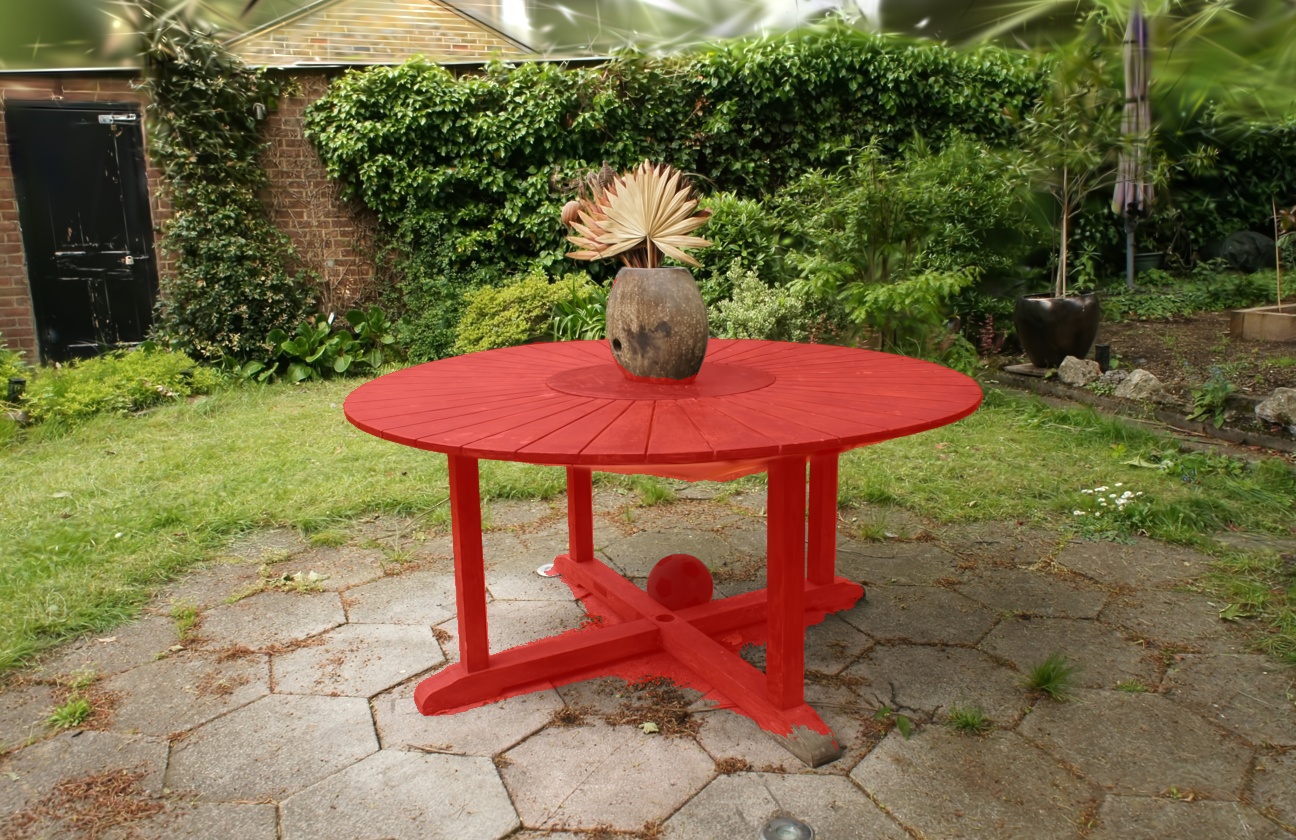}} \\
        \multicolumn{2}{c}{Prompt: Table, Negative Prompt: Vase, Other}\\
        \multicolumn{2}{c}{}\\
        \adjustbox{valign=c}{\includegraphics[width=0.48\linewidth]{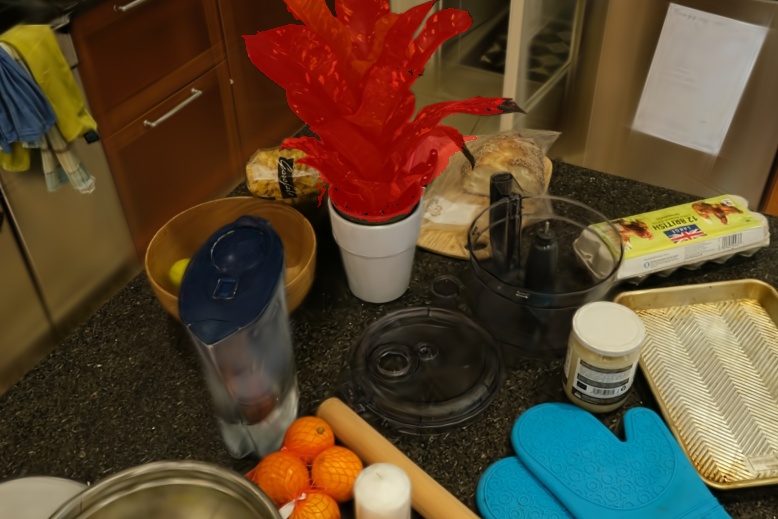}} &
        \adjustbox{valign=c}{\includegraphics[width=0.48\linewidth]{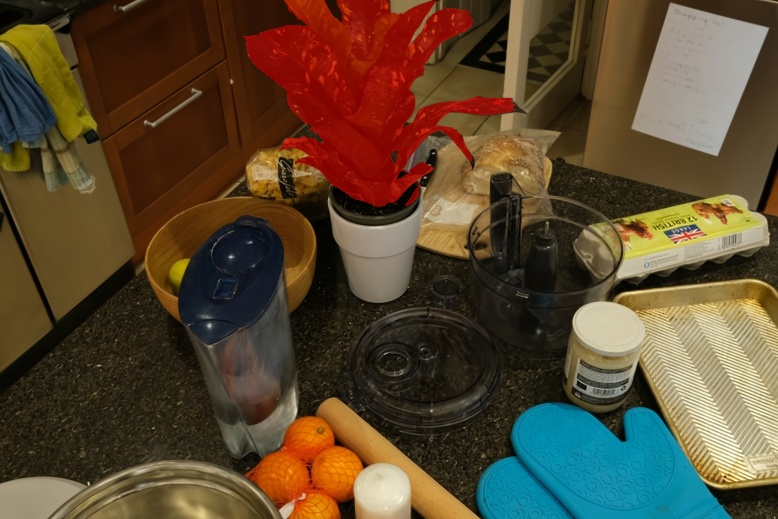}} \\
        \multicolumn{2}{c}{Prompt: Plant, Negative Prompt: Vase, Other}\\
        \multicolumn{2}{c}{}\\
        \adjustbox{valign=c}{\includegraphics[width=0.48\linewidth]{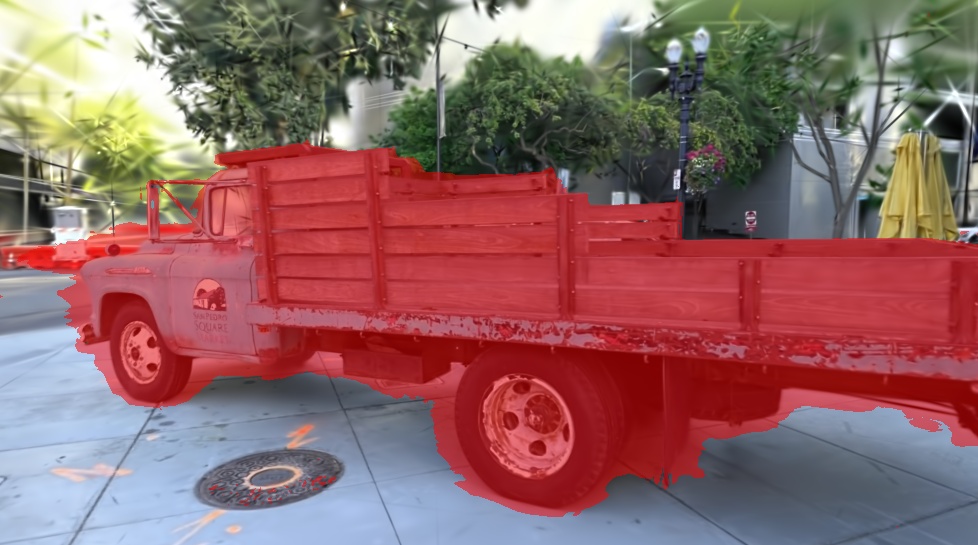}} &
        \adjustbox{valign=c}{\includegraphics[width=0.48\linewidth]{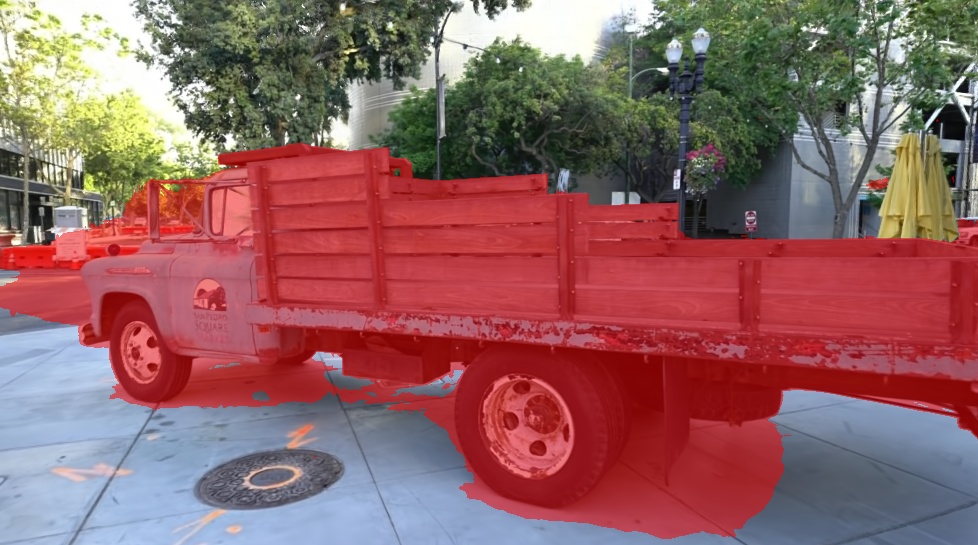}} \\
        \multicolumn{2}{c}{Prompt: Truck, Negative Prompt:  Other}\\
        \multicolumn{2}{c}{}\\
        \bottomrule
    \end{tabular}
    \caption{Comparison of 2D segmentation in Feature 3DGS (F3DGS) and our method. Under each pair of image corresponding positive and negative prompts are given.}
    \label{fig:2d-seg-results}
\end{figure}

\begin{figure*}[h!]
    \centering
    \begin{tabular}{cccc}
        \toprule
        \textbf{Extraction (F3DGS)} & \textbf{Extraction (Ours)} & \textbf{Deletion (F3DGS)} & \textbf{Deletion (Ours)} \\
        \midrule
        \adjustbox{valign=c}{\includegraphics[width=0.24\linewidth]{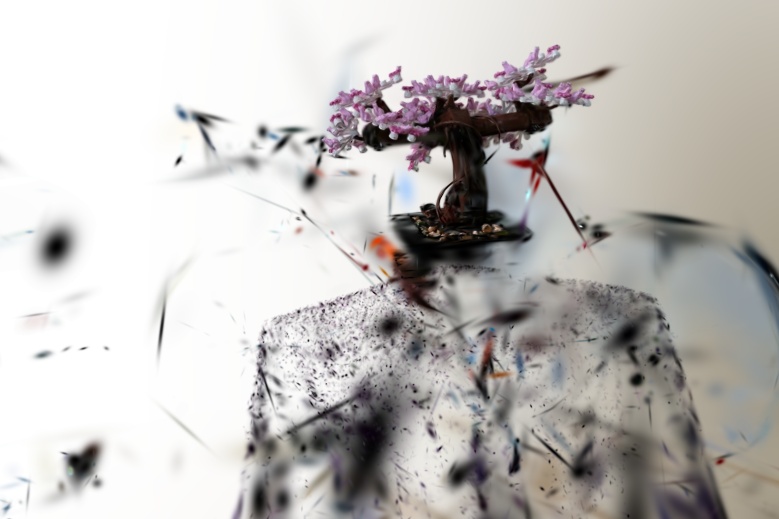}} &
        \adjustbox{valign=c}{\includegraphics[width=0.24\linewidth]{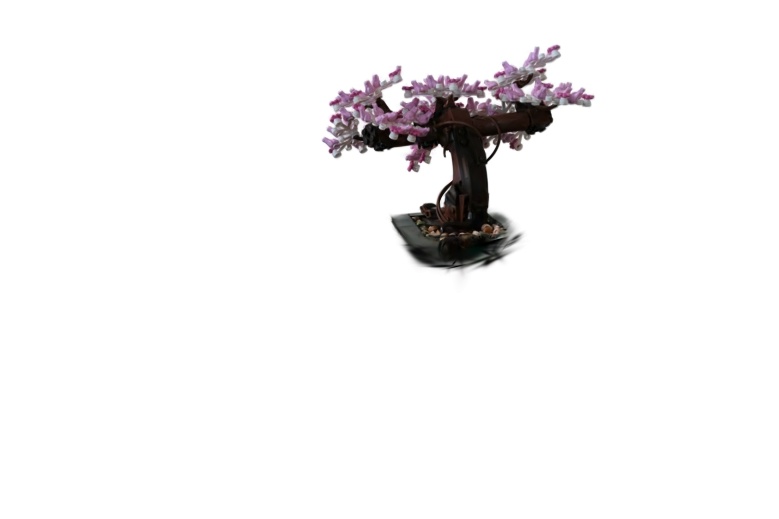}} &
        \adjustbox{valign=c}{\includegraphics[width=0.24\linewidth]{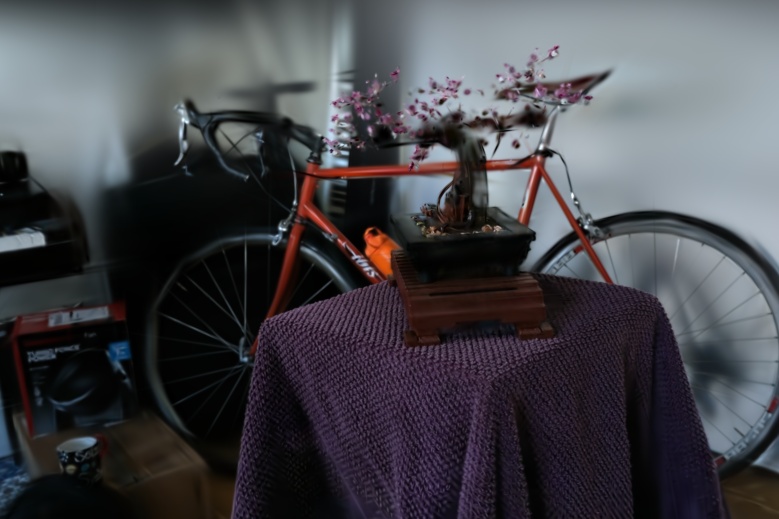}} &
        \adjustbox{valign=c}{\includegraphics[width=0.24\linewidth]{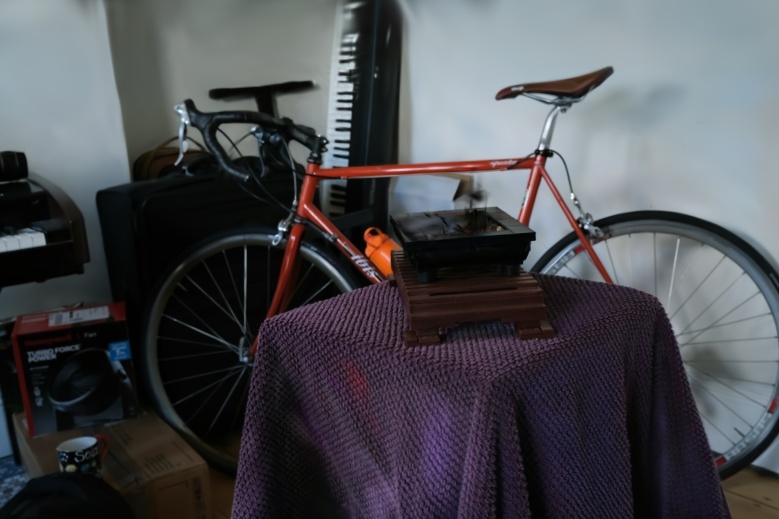}} \\
        \multicolumn{4}{c}{Prompt: Plant, Negative Prompt: Other}\\
        \multicolumn{4}{c}{}\\
        \adjustbox{valign=c}{\includegraphics[width=0.24\linewidth]{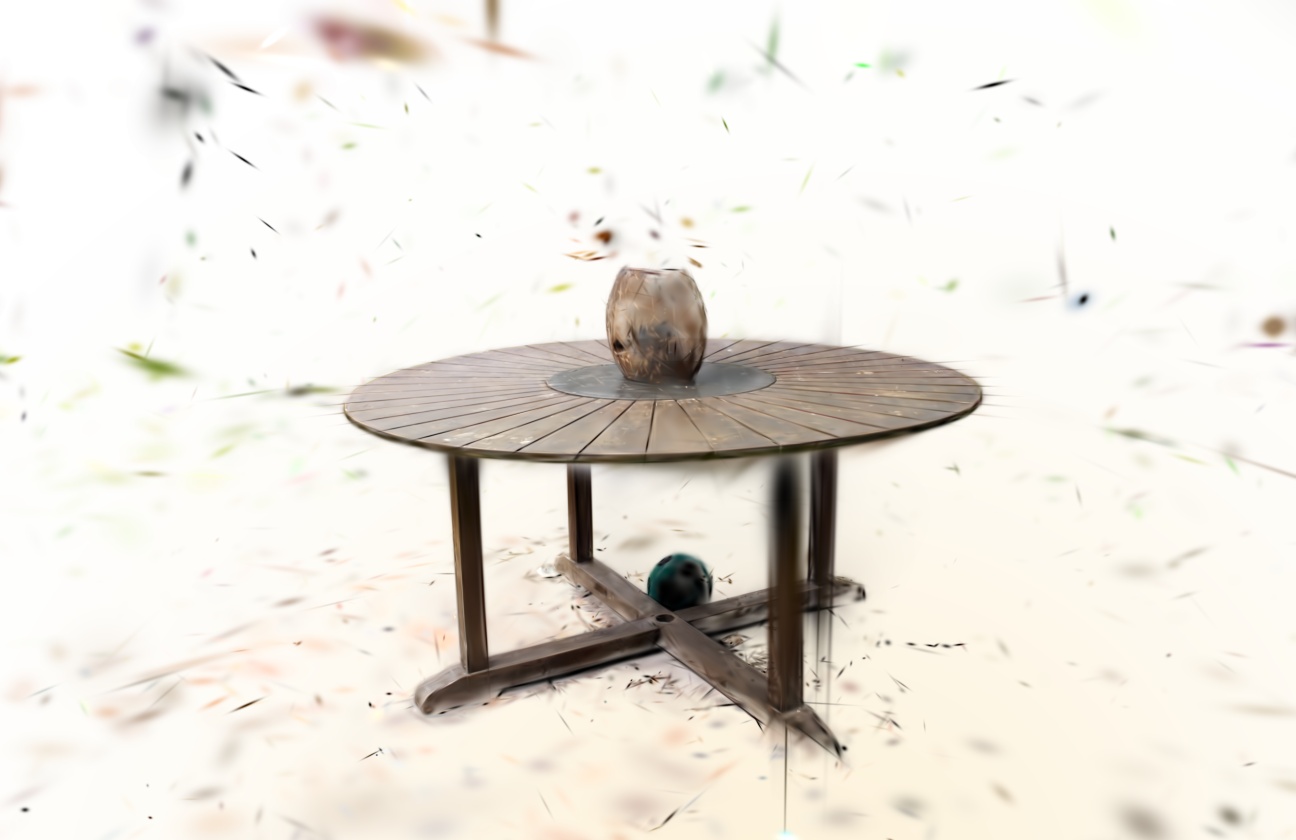}} &
        \adjustbox{valign=c}{\includegraphics[width=0.24\linewidth]{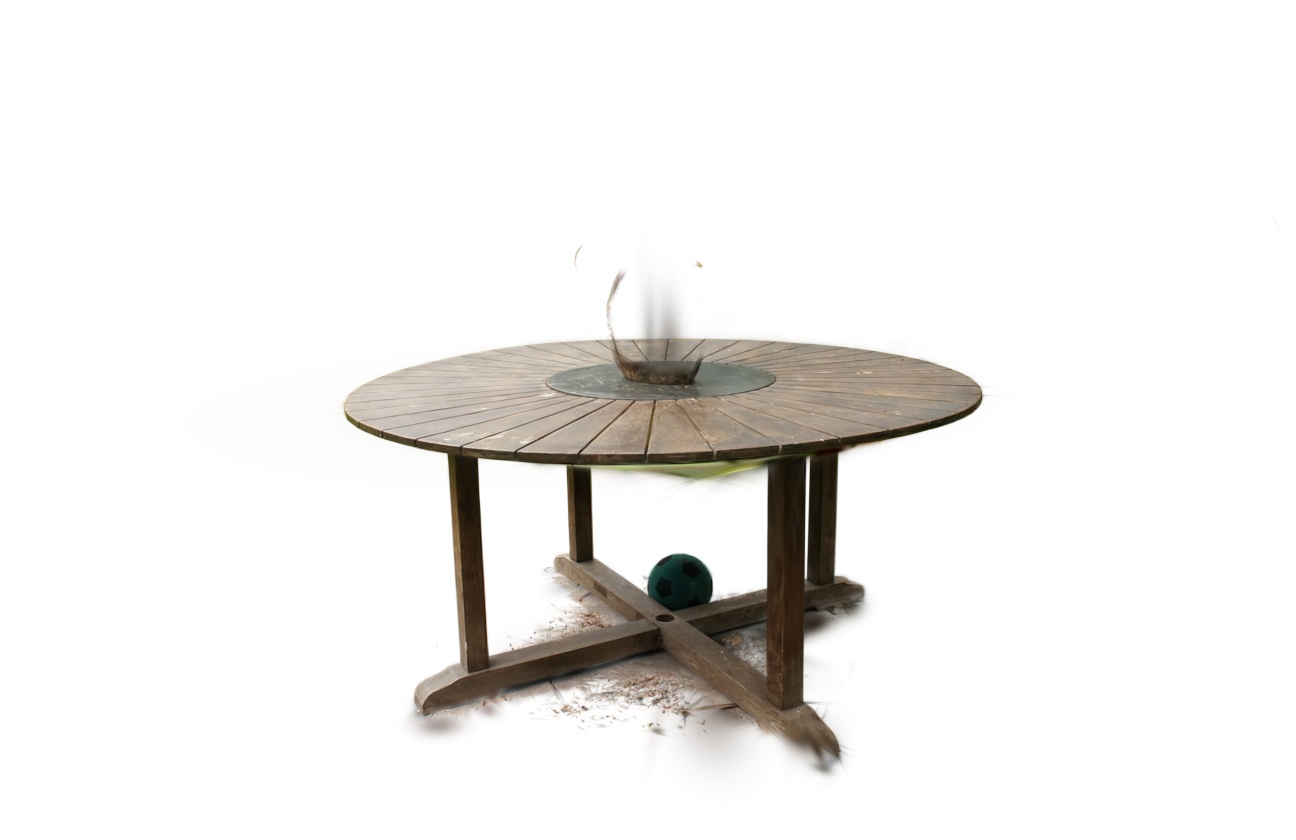}} &
        \adjustbox{valign=c}{\includegraphics[width=0.24\linewidth]{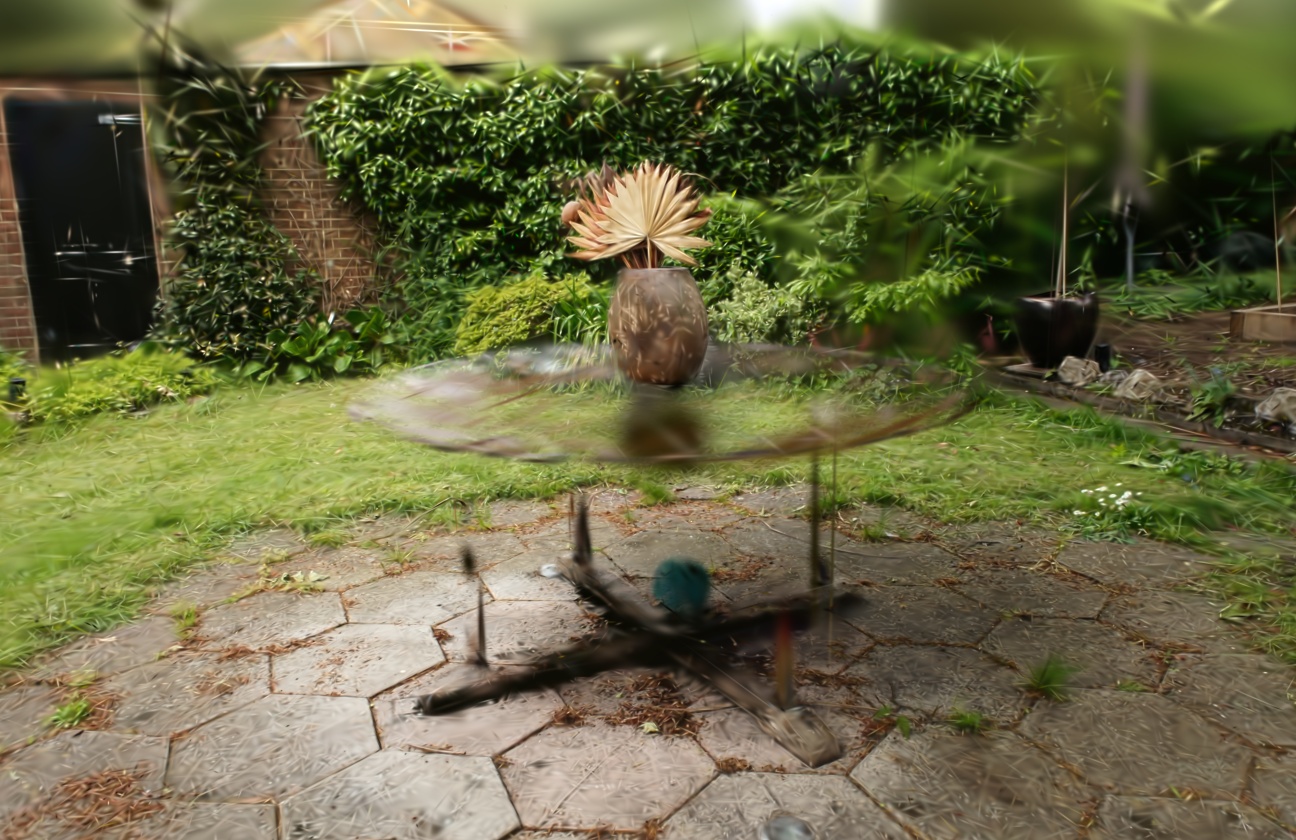}} &
        \adjustbox{valign=c}{\includegraphics[width=0.24\linewidth]{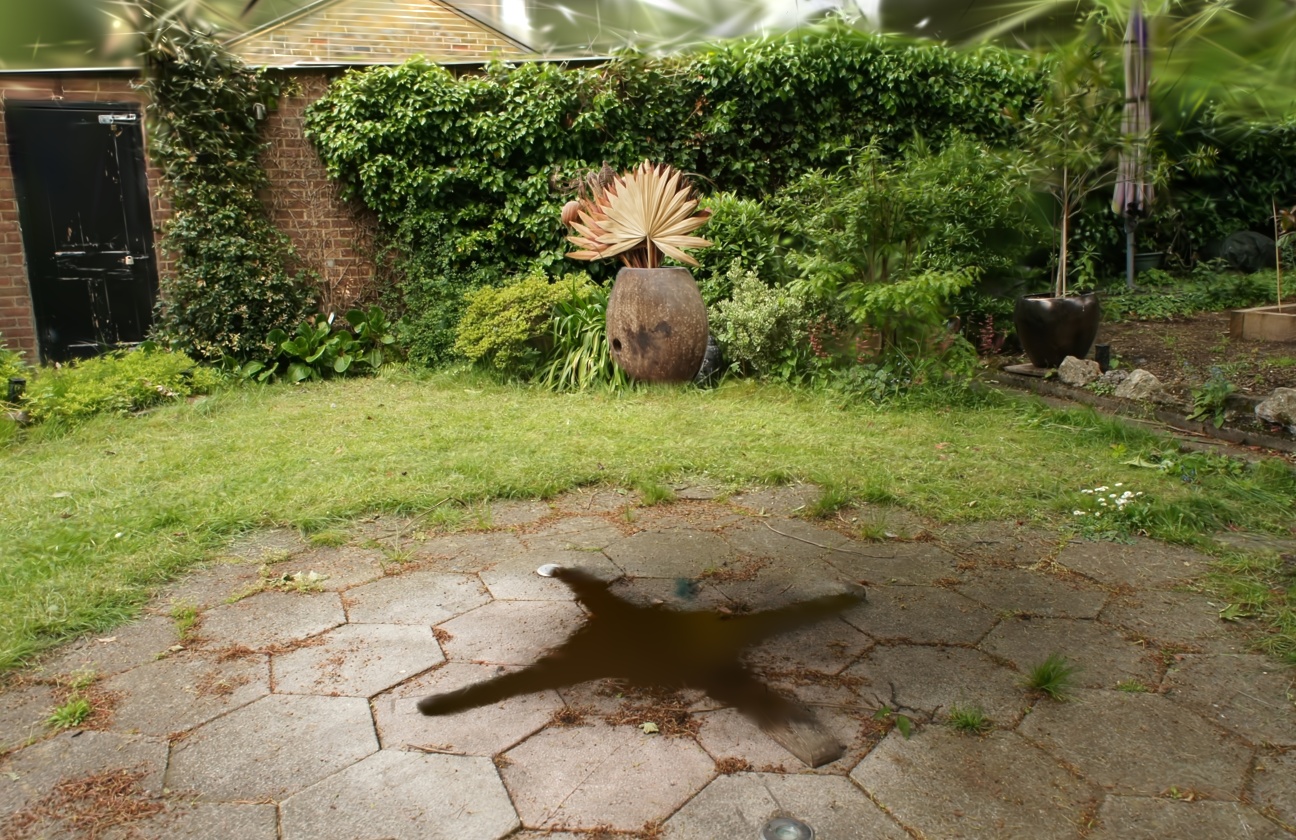}} \\
        \multicolumn{4}{c}{Prompt: Table, Negative Prompt: Vase, Other}\\
        \multicolumn{4}{c}{}\\
        \adjustbox{valign=c}{\includegraphics[width=0.24\linewidth]{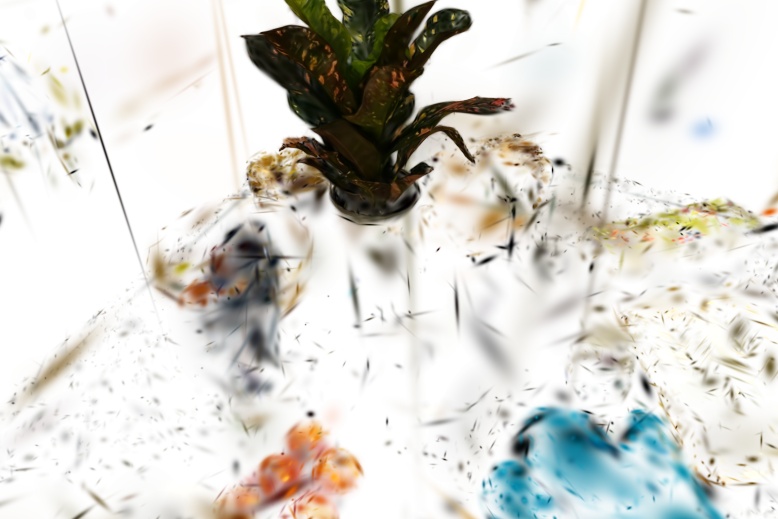}} &
        \adjustbox{valign=c}{\includegraphics[width=0.24\linewidth]{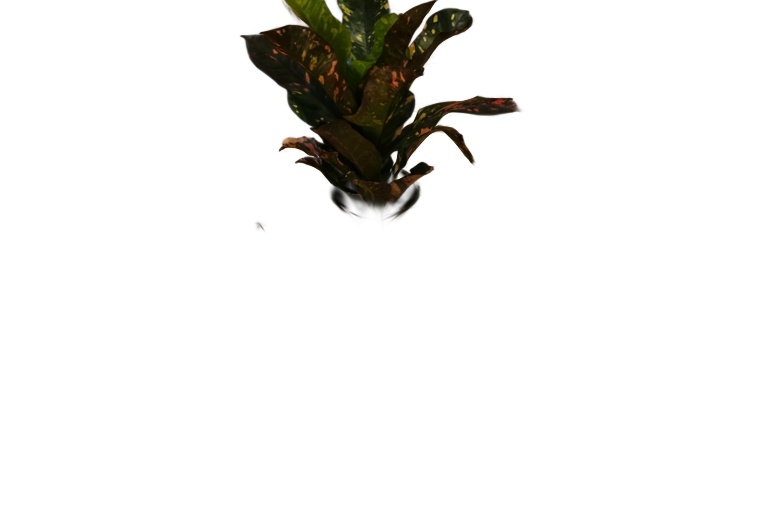}} &
        \adjustbox{valign=c}{\includegraphics[width=0.24\linewidth]{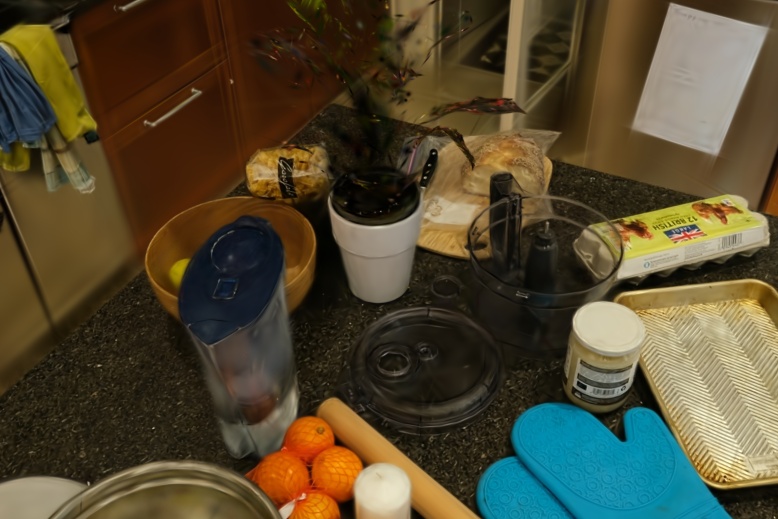}} &
        \adjustbox{valign=c}{\includegraphics[width=0.24\linewidth]{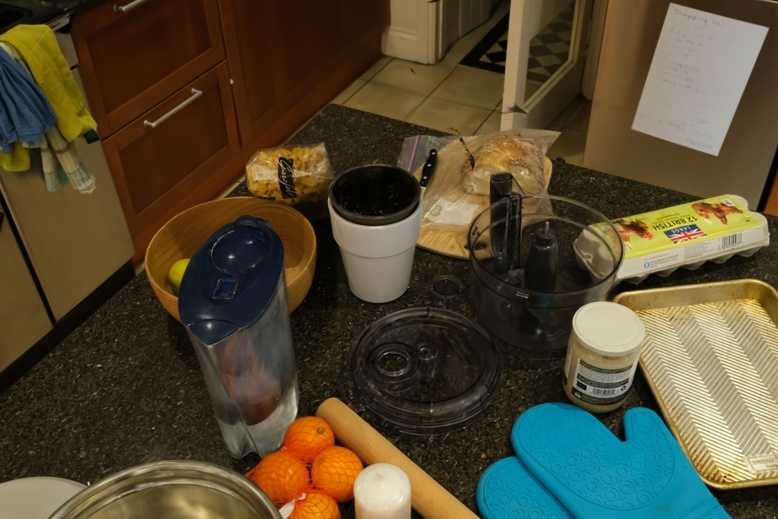}} \\
        \multicolumn{4}{c}{Prompt: Plant, Negative Prompt: Vase, Other}\\
        \multicolumn{4}{c}{}\\
        \adjustbox{valign=c}{\includegraphics[width=0.24\linewidth]{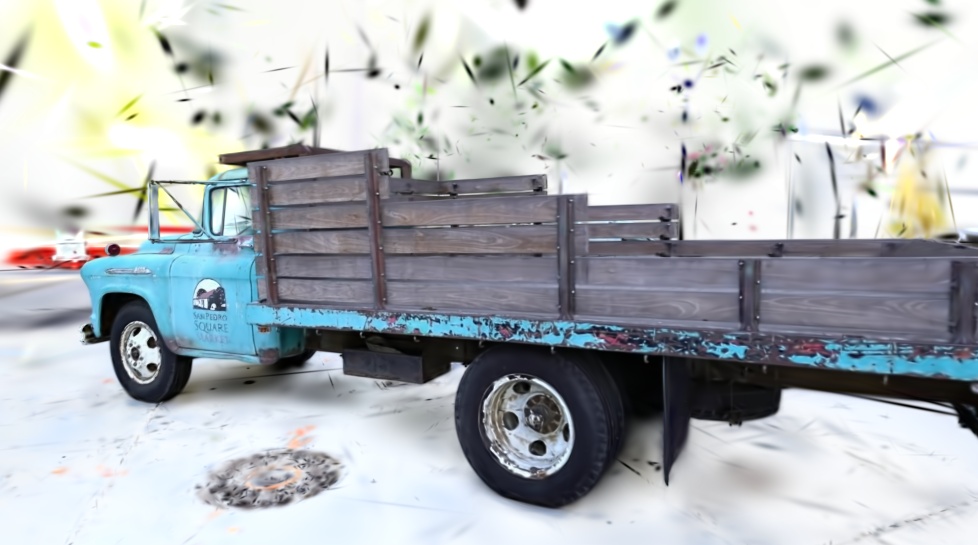}} &
        \adjustbox{valign=c}{\includegraphics[width=0.24\linewidth]{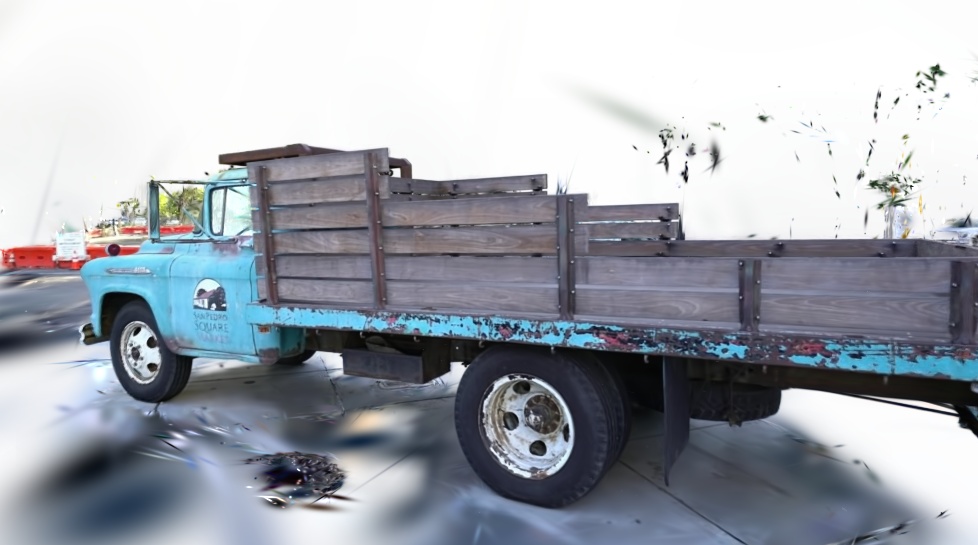}} &
        \adjustbox{valign=c}{\includegraphics[width=0.24\linewidth]{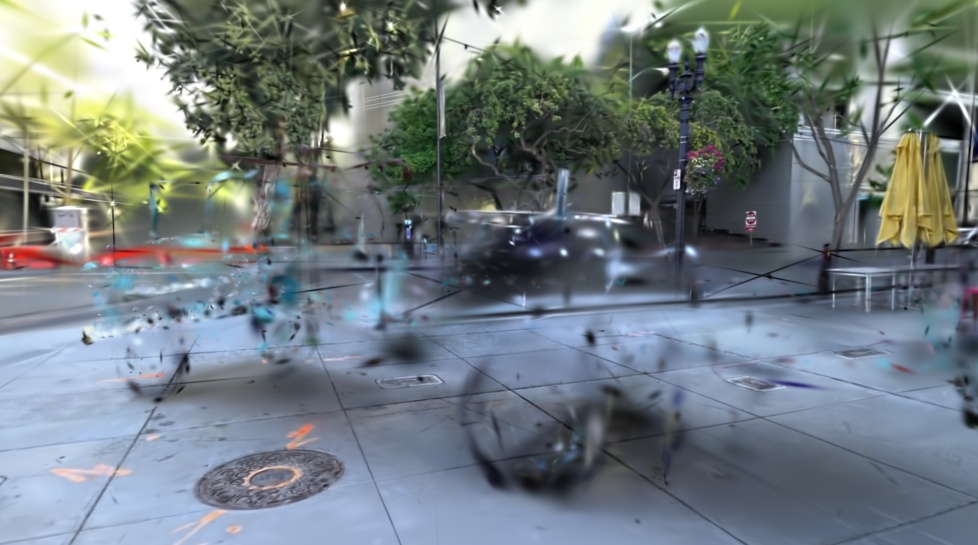}} &
        \adjustbox{valign=c}{\includegraphics[width=0.24\linewidth]{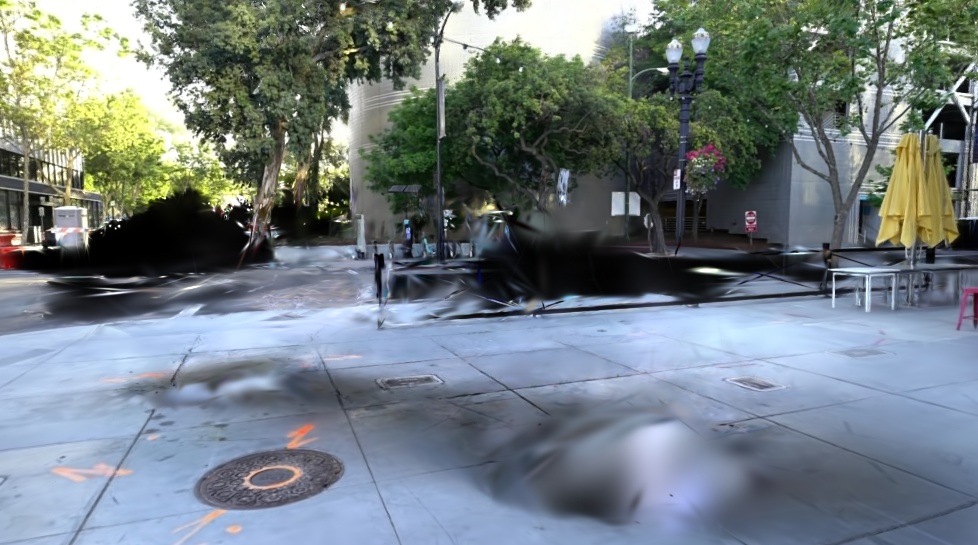}} \\
        \multicolumn{4}{c}{Prompt: Truck, Negative Prompt:  Other}\\
        \multicolumn{4}{c}{}\\
        \bottomrule
    \end{tabular}
    \caption{Comparison of 3D segmentation in Feature 3DGS (F3DGS) and our method. Under each pair of image corresponding positive and negative prompts are given. Clearly our method produces less outliers.}
    \label{fig:3d-seg-results}
\end{figure*}

The qualitative 2D segmentation results are given in figure~\ref{fig:2d-seg-results}, and 3D segmentation results are given in figure~\ref{fig:3d-seg-results}. For 2D segmentation, both methods produce similar results. But for the 3D segmentation, our method gives better results.

The feature back-projection is completed in a single pass through the training views, with the entire process—including ground truth feature map generation—taking an average of 2-3 minutes. During inference, our method achieves object segmentation in just 30 ms, including both text encoding and similarity calculation. This represents a 900x speedup compared to segmentation using masked gradients \cite{joji2024gradient}, a method designed for clean segmentation, when considering the full pipeline including mask generation. In our approach, the majority of the latency arises from text encoding, while similarity calculation is highly efficient, requiring less than 1 ms.

\subsection{Affordance Transfer}

We transfer DINOv2 features to each scene of the \cite{Myers:ICRA15} dataset. Then use source images and annotations from \cite{joji2024gradient} and transfer it directly to the Gaussians of target scene. We label the method from \cite{joji2024gradient} as 2D-2D-3D transfer because source annotations are transferred to 2D target frames before applying to 3D and our method as 2D-3D for contrasting.

\begin{figure}
    \centering
    \begin{subfigure}[b]{0.4\linewidth}
    \includegraphics[width=\linewidth]{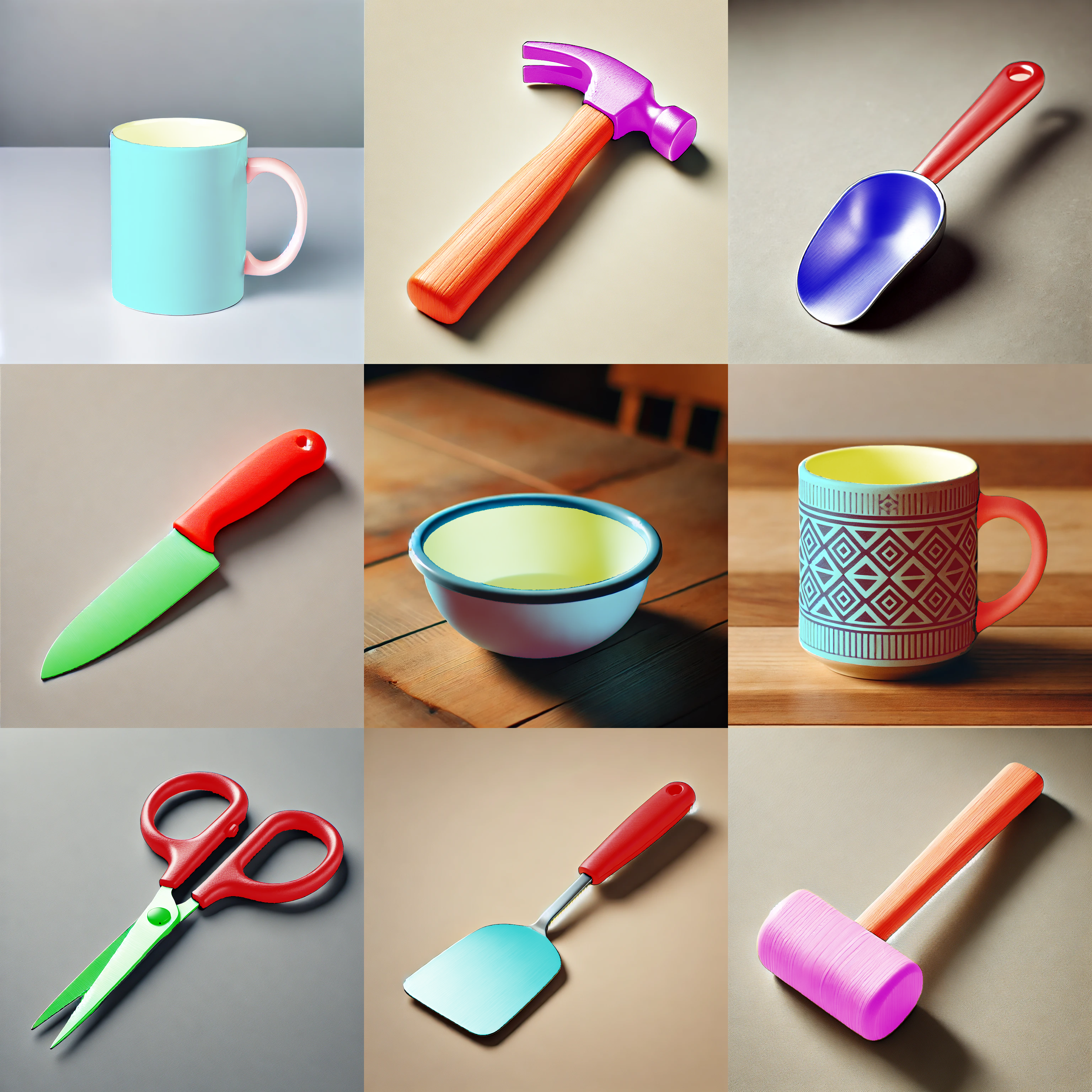}
    \caption{Source Images with annotations}
    \end{subfigure} %
    \begin{subfigure}[b]{0.5\linewidth}
    \includegraphics[width=\linewidth,height=0.8\linewidth]{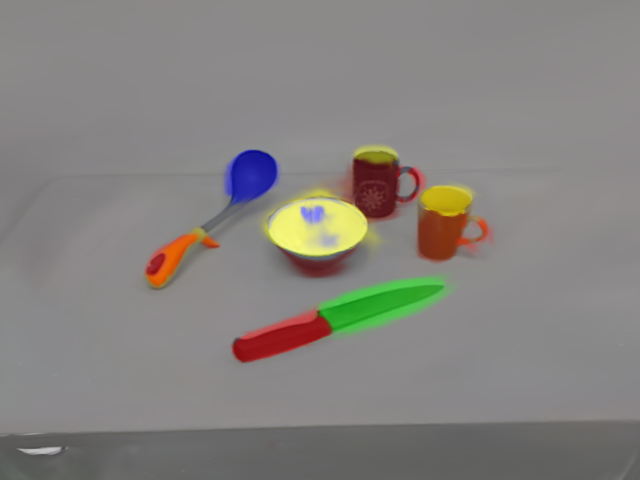}
    \caption{Target scene with transferred annotations}
    \end{subfigure} %
    \caption{Qualitative result of affordance transfer. The example images are different but the same category as that of target scene.}
    \label{fig:affordance-transfer}
\end{figure}

\begin{table}[htbp!]
{\scriptsize
\caption{Table showing comparison affordance transfer with masked gradients and our method. The label 2D-3D indicate our method.}
    \label{tab:affordance-quantitative}
    \centering
    \begin{tabular}{lcccccc}
        \toprule
        & \multicolumn{2}{c}{\textbf{mIoU $\uparrow$}} & \multicolumn{2}{c}{\textbf{Recall $\uparrow$}} & \multicolumn{2}{c}{\textbf{Time Taken (s) $\downarrow$}}\\
        \cmidrule(r){2-3} \cmidrule(l){4-5} \cmidrule(l){6-7}
        \textbf{Scene} & \textbf{2D-2D-3D} & \textbf{2D-3D } & \textbf{2D-2D-3D} & \textbf{2D-3D } & \textbf{2D-2D-3D} & \textbf{2D-3D } \\
       \midrule
       1 & \textbf{47.87} & 42.80 & \textbf{67.77} & 67.11 & 293.88 & \textbf{5.22}\\
       2 & \textbf{55.63} & 53.28 & 81.07 & \textbf{82.55} & 317.12 & \textbf{8.03}\\
       3 & \textbf{60.50} & 57.82 & \textbf{86.95} & 86.68 & 142.79 & \textbf{7.58}\\
       \midrule
        Mean & \textbf{54.67} & 51.30   & 78.60 & \textbf{78.78} & 251.26 &  \textbf{6.94}\\
       \bottomrule
    \end{tabular}}
\end{table}

\subsection{Identity Encoding}

\begin{figure}
    \centering
    \begin{subfigure}[b]{0.45\linewidth}
    \includegraphics[width=\linewidth]{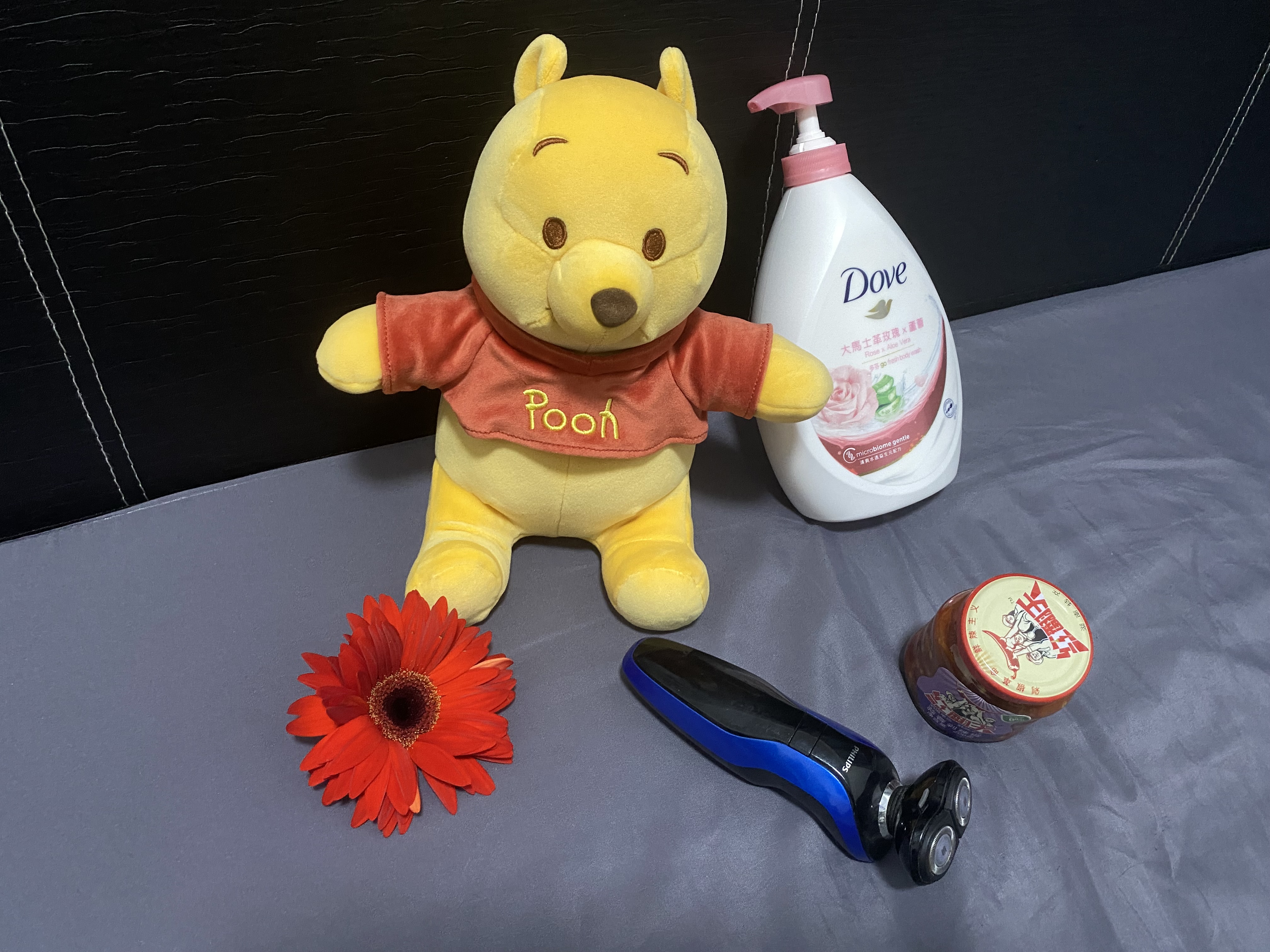}
    \caption{Original View}
    \end{subfigure} %
    \begin{subfigure}[b]{0.45\linewidth}
    \includegraphics[width=\linewidth]{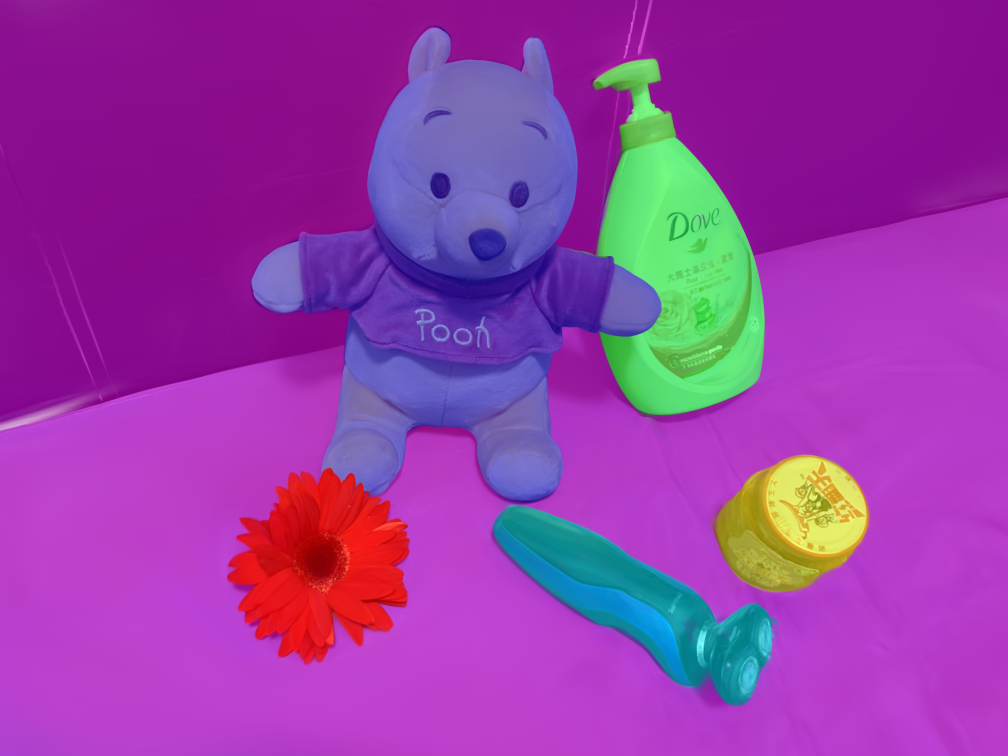}
    \caption{Rendering with grouping}
    \end{subfigure} %
    \begin{subfigure}[b]{0.45\linewidth}
    \includegraphics[width=\linewidth]{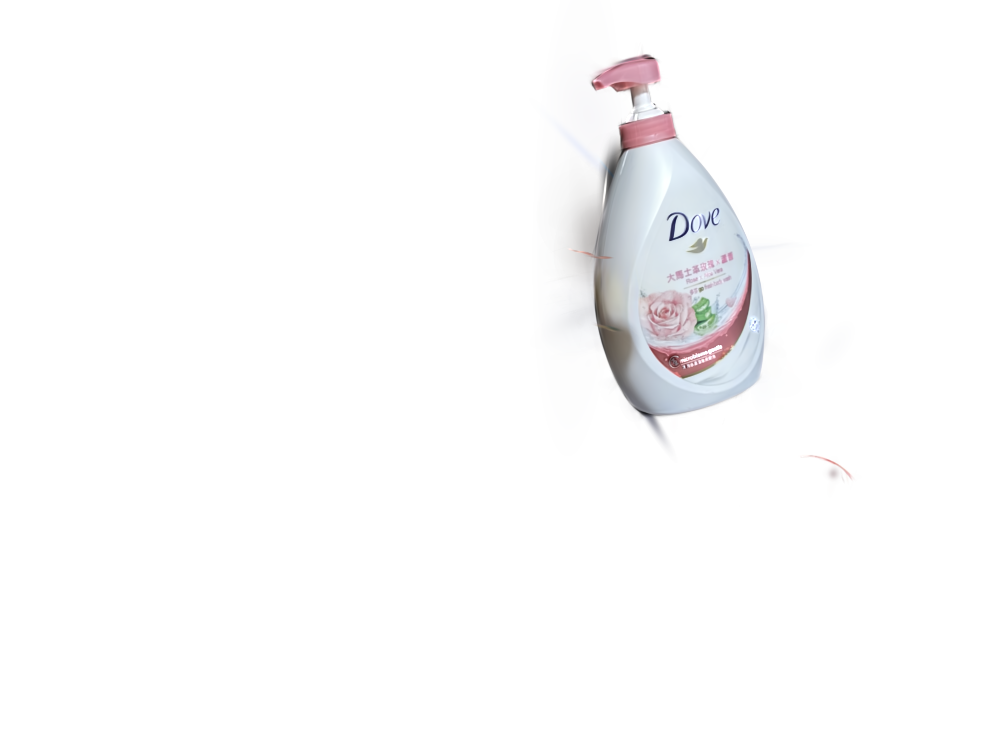}
    \caption{Extraction}
    \end{subfigure} %
    \begin{subfigure}[b]{0.45\linewidth}
    \includegraphics[width=\linewidth]{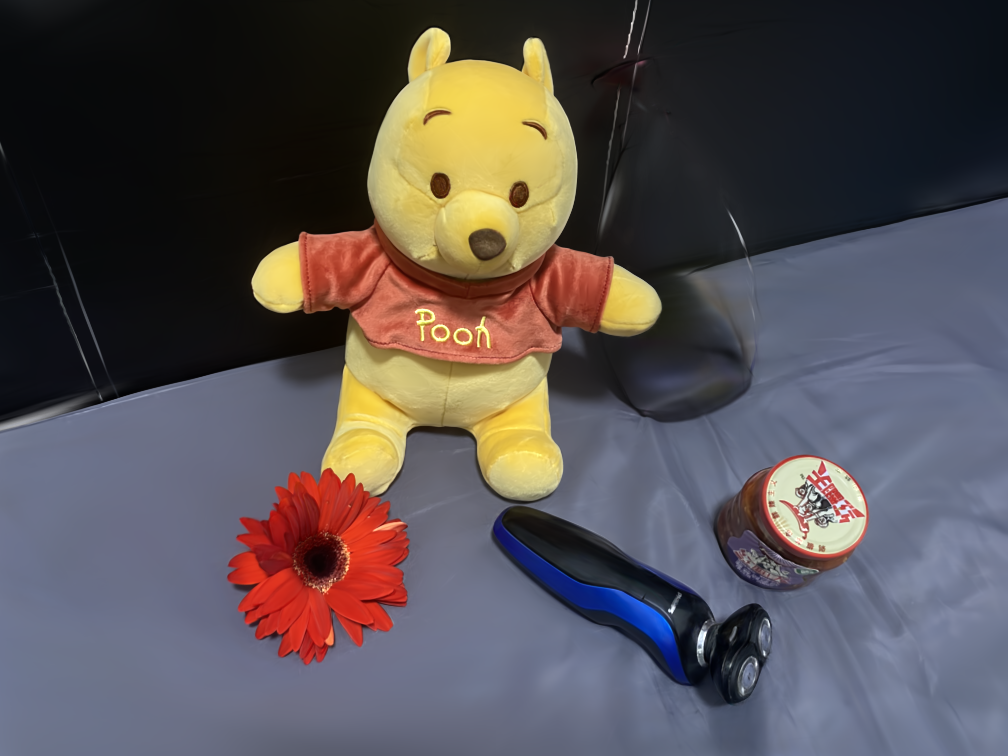}
    \caption{Deletion}
    \end{subfigure} %
    \caption{Sample result of orthogonal identity encoding. We use only about 5 frames with ground truth. The scene is trained with 30 imags. We are able to extract the occluded object as well as delete it. Note that the results are shown without post processing to remove outliers.}
    \label{fig:identity-encoding}
\end{figure}

\textbf{Orthogonal Encoding}: For simplicity we use one-hot encoding for identity encoding. That is one element is 1 and rest of them are 0. The qualitative results are shown in figure \ref{fig:identity-encoding}. The scene is taken from 3D-OVS dataset\cite{liu2023weakly}.

\textbf{Contrastive Encoding}: We use LERF-Mask dataset introduced for the identity encoding method Gaussian Grouping\cite{gaussian_grouping} for quantitative evaluation. Here we use 16 dimension embeddings for each group. Total number of groups are around 200. 

We first train a classifier to predict the group over the embeddings. We make sure to use a contrastive loss to make embeddings far apart from each other. Then back-project this embeddings to each Gaussian. We use the classifier to predict the groups each rendered pixel belongs.

We follow the evaluation protocol from \cite{gaussian_grouping} to evaluate our method. See table~\ref{tab:grouping-quantitative} for the quantitative evaluation. Note that we evaluate using 2D mIoU. Surprisingly we are able to get similar performance as that of Gaussian Grouping which trains identity encoding along with each scene. Our method takes only 10 seconds for training the classifier and 10 seconds to transfer the identity encoding.

\begin{table}[h]
    \centering
    {\footnotesize
    \caption{Comparison of mIoU scores for different models across Figurines, Ramen, and Teatime datasets. Bold indicates best performance and underline indicates second best. Our method is comparable to the state of the art Gaussian Grouping\cite{gaussian_grouping}}. The readings other that ours are from \cite{gaussian_grouping}.
    \label{tab:grouping-quantitative}
    \begin{tabular}{@{}lcccc@{}}
        \toprule
        \textbf{Model} & \textbf{Figurines}  & \textbf{Ramen}  & \textbf{Teatime} & \textbf{Mean} \\ 
        \midrule
        DEVA \cite{cheng2023tracking}           & 46.2 & 56.8 & 54.3 & 52.4\\
        LERF \cite{lerf2023}          & 33.5 & 28.3 & 49.7 & 37.2\\
        SA3D \cite{cen2023segment}          & 24.9 & 7.4  & 42.5 & 24.9\\
        LangSplat \cite{qin2023langsplat}     & 52.8 & 50.4 & {69.5} & 57.6\\
        \midrule
        Gaussian Grouping \cite{gaussian_grouping}  & \underline{69.7} & \textbf{77.0} & \underline{71.7} & \underline{72.8}\\
        Ours  & \textbf{73.5} & \underline{72.7} & \textbf{74.1} & \textbf{73.4}\\
        \bottomrule
    \end{tabular}}
\end{table}

\section{Discussion and Conclusion}
\label{sec:discussion}
We have introduced a novel, training-free, efficient, scalable alternative to feature field distillation in Gaussian splatting. Our method achieves fast, clean segmentation by directly querying the features associated with each Gaussian.

Our approach aggregates features from all available training views in a single pass, mitigating any inconsistencies in individual 2D feature maps through an averaging effect. Nevertheless, minor imperfections may still propagate, though they have minimal impact on overall performance.

Since our method does not update Gaussian parameters, it does not benefit from the regularization effect inherent in traditional feature field methods. However, we found no significant drawbacks when compared to feature field methods—in fact, our approach is considerably faster for generating feature embeddings for Gaussians and produces superior qualitative results.

{
    \small
    \bibliographystyle{ieeenat_fullname}
    \bibliography{main}
}

\end{document}